# Learning Based High-Level Decision Making for Abortable Overtaking in Autonomous Vehicles

Ehsan Malayjerdi, Gokhan Alcan, Eshagh Kargar,
Hatem Darweesh, Raivo Sell, Ville Kyrki *Senior Member, IEEE*

*Abstract*—Autonomous driving is a growing technology that aims to enhance safety, accessibility, efficiency, and convenience through autonomous maneuvers ranging from lane change to overtaking. Overtaking is one of the most challenging maneuvers for autonomous vehicles, and current techniques for autonomous overtaking are limited to simple situations. This paper studies how to increase safety in autonomous overtaking by allowing the maneuver to be aborted. We propose a decision-making process based on a deep Q-Network to determine if and when the overtaking maneuver needs to be aborted. The proposed algorithm is empirically evaluated in simulation with varying traffic situations, indicating that the proposed method decreases the collision rate and improves the overtaking maneuver safety by aborting the overtaking decision if needed. Furthermore, the approach is demonstrated in real-world experiments using the autonomous shuttle iseAuto.

## I. INTRODUCTION

AUTONOMOUS driving is becoming increasingly popular as a research topic, as the level of autonomy is moving from driver assistance to autonomous operation. Current autonomous vehicles (AV) aim to tackle an increasing complexity of maneuvers, including lane-following, lane-changing, merging, and overtaking [1], [2].

Overtaking is one of the hardest problems to solve as it is a highly risky maneuver and causes a lot of accidents [3]. During an overtaking maneuver, a driver needs to predict how a traffic situation develops over time in order to estimate whether a sufficient time window is available for overtaking. This is difficult also for human drivers, especially in complex or high-speed situations. Because of its inherent complexity, overtaking is influenced by a number of factors, including road and traffic conditions [4], and type of overtaking (flying overtaking, accelerative overtaking, or piggybacking) [5], [6].

Autonomous overtaking is usually considered from two perspectives, how to overtake (choice of trajectory) and when to overtake (decision making). The choice of overtaking trajectory is made using either simple heuristics (e.g. two lane changes) or with a local path planner. Planning algorithms have been successfully applied to autonomous overtaking [7], but a few issues remain. In particular, it is challenging to ensure safety under incomplete information of factors including the current and especially future traffic situation, interactions with other road users (cars and pedestrians), weather conditions, traffic rules, and road quality [4], [8].

Due to the fact that the future is unpredictable, online reactive decision-making is beneficial in addition to planning. Decision-making in uncertain and dynamic traffic situations is a general challenge in autonomous driving [9], [10]. For overtaking, researchers have proposed a variety of approaches, including fuzzy logic [11], rule-based methods [12], and learning-based approaches [13]–[16]. However, the existing approaches are limited to relatively simple situations. For rule-based and fuzzy approaches this is usually explicit in that overtaking is only allowed in very limited situations, while for learning-based methods, current works have also considered simple scenarios evaluated in simulation. Furthermore, few existing works address the fact that after an overtaking maneuver has been initiated, new information may be gained, for example, an oncoming vehicle is detected during the maneuver, in which case the maneuver might need to be aborted.

We propose a high-level learning-based decision-making framework for autonomous overtaking, which decreases the collision rate compared to rule-based methods. The main contributions of our work are:

1) A deep Q-Network (DQN) based high-level decision-making mechanism with "*Following*", "*Overtaking*" and "*Aborting*" actions that lowers crash rate.
2) Compatible integration of learning-based decision-making with an existing path and trajectory planning using OpenPlanner.
3) Simulation studies for varying traffic conditions demonstrating that the proposed architecture collision rate by aborting the overtaking if needed.
4) Validation via implementation on the Tallinn University of Technology (TalTech) automated shuttle - iseAuto [17], [18] shows that the proposed method can be deployed in such systems without changing the existing path and trajectory planners.

The organization of this paper is as follows: After reviewing related work in Sec. II, the proposed architecture integrating learning-based high-level decision making and planning is described in Sec. III. Simulation experiments with statis-

This work was supported by the Academy of Finland under Grants 328399 and 345661.

E. Malayjerdi and R. Sell are with Department of Mechanical and Industrial Engineering, Tallinn University of Technology, Tallinn, Estonia, `firstname.lastname@taltech.ee`.
G. Alcan, E. Kargar and V. Kyrki are with Department of Electrical Engineering and Automation, Aalto University, Helsinki, Finland, `firstname.lastname@aalto.fi`.
H. Darweesh is with Graduate School of Informatics, Nagoya University, Nagoya, Japan, `hatem.darweesh@g.sp.m.is.nagoya-u.ac.jp`.

tical analysis and real-world implementation are presented in Sec. IV and discussed in Sec. V. Finally, we provide conclusive remarks in Sec. VI.

## II. Related Work

AVs compute overtaking trajectories through decision-making algorithms in their planning modules. In this module, the most efficient path for overtaking is selected by path planning, and then this path is used for driving with trajectory planning. Decision-making or behavior selector is extremely important and is considered to be the human brain [19]. Based on this decision, the automated vehicle will either overtake or abort the overtaking process.

In the field of path planning for overtaking in AVs, [7] proposed an optimized sigmoid-based path planning algorithm that guarantees a smooth, fast and safe overtaking maneuver. Although the overtaking method has shown promising results in simulations and experiments, it is limited to simple overtaking of a stopped vehicle with no oncoming vehicles on the opposite lane. Bing Lu et al [20] proposed an adaptive potential field-based path planning for complex autonomous driving scenarios such as overtaking maneuvers. Although the proposed method demonstrates a robust and smooth overtaking performance, the method lacks a safety check, and the simple simulation without software or vehicle in the loop was not reliable enough. Based on analyzing human drivers in path planning and utilizing artificial potential field functions, [21] proposes a method for autonomous overtaking. The proposed approach is found to be effective in overtaking both static and moving obstacles. However, In this study, overtaking is simplified by using a constant-speed vehicle without any oncoming traffic.

The trajectory planning layer computes a safe, comfortable, and dynamically optimal trajectory from the vehicle's current position and configuration to the goal configuration [22]. Different studies focus on trajectory planning modules to solve the overtaking maneuver issues in AVs. A trajectory planning algorithm proposed by Shilp et al. [23] for AVs overtaking maneuvers on highways, in a combination of the potential field function and model prediction control. [24] proposed combining artificial potential fields and MPC to develop a trajectory planning controller for obstacle avoidance in high-speed overtaking. An optimized multi-objective trajectory planning for AVs is proposed in [25]. The particle filtering proposed in [26] solves the trajectory planning problem as a nonlinear non-Gaussian estimation. These methods have shown smooth and robust overtaking in simulation. Nevertheless, the low fidelity simulation was not reliable enough, and without sensor data for perception there was no safety verification.

Many studies have been conducted to address the overtaking decision-making problem. A human-like decision-making platform based on Nash equilibrium and Stackelberg game theory for overtaking maneuvers in AVs proposed in [27]. However, despite showing a great ability to address the decision-making problem of AVs, it is limited by computation resources and its inability to handle complex scenarios. [28] proposes a fuzzy logic controller for overtaking maneuvers. The fuzzy controller in this method acts as a driver in overtaking situations. Although the proposed method achieved reasonable results in their three experiments, it may lack performance when the GPS (Global Positioning System) signal is lost or not reliable. A finite state machine as a high-level decision-making and chance-constrained model predictive control as a trajectory planning is proposed in [12] to ensure safety, optimize efficiency, and passenger comfort in overtaking maneuvers. The proposed method has shown safe and comfortable overtaking performance in comparison with rule-based methods. In spite of this, the simulation did not provide adequate reliability, and the method did not provide safety verification.

Autonomous overtaking has also been addressed as a reinforcement learning problem. The proposed method in [29] addresses the training of overtaking decision-making policies by the Q-learning algorithm. Overtaking policies are determined by a Q-learning algorithm and evaluated in several traffic conditions based on two main indicators, the average velocity and the minimum distance between vehicles. This study was carried out without any experiments and merely based on a simple simulation that is insufficient as a verification testbed, although it has shown that it can generate appropriate policies under different traffic situations. A multi-objective approximate policy iteration (MO-API) algorithm to learn policies for overtaking maneuvers modeled by a Markov decision process with multiple goals developed in [30]. However, it is important to note that while simulations and experiments suggest that the proposed RL approach can determine the appropriate overtaking policies in different traffic conditions, the overtaking scenario takes place on the highway without any traffic appearing in the opposite direction. To learn policies for overtaking decisions, a Double Deep Q-learning (Double DQN) agent was proposed in [31]. As a result, with Double DQN as a learning method, continuous states can directly be used to learn a more precise policy without discretization, which is insufficient in this study since a simple simulation was used instead of experiments. A novel hierarchical reinforcement learning based on the semi-Markov decision process and motion primitives for overtaking is proposed in [32]. In this method, high-level decision-making is combined with low-level control by motion primitives. The proposed method has demonstrated precise control in overtaking maneuvers. However, the simulation was simple and lack of safety verification.

Several studies focused on integrated planner which is responsible for planning and decision making. OpenPlanner [33], [34] is an integrated open-source planner which based on a vector (road network) map, it generates global reference paths. Following this global path, the local planner calculates a local trajectory that is obstacle-free by using sampled roll-outs. Various factors, such as collisions, traffic rules, transitions, and distance from the center, are considered in selecting the safe trajectory. A behavior generator acts as a decision maker by calculating probability values for other vehicles' intentions and trajectories, and an intention estimator calculates probabilities for other vehicles' intentions and trajectories. Baidu Apollo [35] is an open-source integrated planner based on Apollo [36]. Safety, comfort, and scalability are considered in developing the proposed integrated planner.

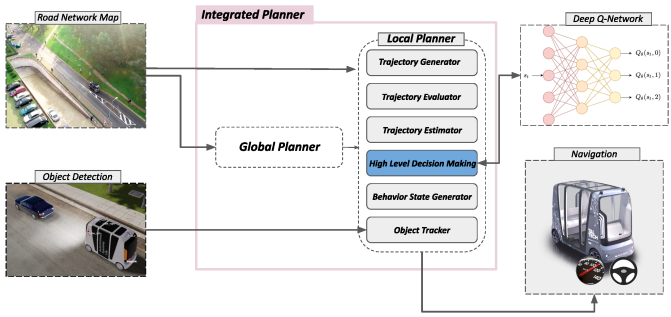

Fig. 1. Proposed learning based autonomous overtaking architecture

However, none of the works above address the issue of aborting the maneuver, which has only been discussed in the literature from the planning perspective [37]. The work focuses on trajectory planning using model predictive control, while a hand-crafted finite state machine is used to choose whether the maneuver is aborted, which is the primary problem addressed in this work.

We believe that safety is a key issue in overtaking maneuvers that must be addressed in order to minimize risks. In this paper, a high-level decision-making algorithm based on open source integrated planner and reinforcement learning proposed in order to generate fast, smooth, safe, and reliable overtaking maneuvers in complex scenarios. Here we propose an experimental evaluation of the overtaking maneuver using an autonomous shuttle in an urban environment as opposed to previous literature that studied the maneuver only in simulation.

## III. PROPOSED INTEGRATED ARCHITECTURE

Proposed method targets to replace the rule-based high-level decision-making mechanisms of integrated planners with a Reinforcement Learning Agent (RL-Agent) (Fig 1). In the following subsections, learning-based decision-making of abortable overtaking and its integration into the local planner of an existing integrated planner will be detailed.

### A. Learning Based Decision Making

High-level decision making block of the local planner is responsible to choose one of the three actions from *following*, *overtaking*, and *aborting* based on the road and traffic conditions (Fig 2). Those actions are then employed in the local planner by setting the following parameters appropriately:

- Velocity: The velocity of autonomous vehicle in meters per second (m/s)
- Acceleration : The autonomous Vehicles' ability to accelerate.
- Following Distance: The minimum distance between the autonomous vehicle and the vehicle in front is specified by this parameter.
- Avoiding Distance: To avoid an accident with a vehicle in front, a minimum distance must be maintained.
- Rollout Number: The number of potential routes generated by the local planner.
- Rollout Id: The selected rollout by local planner for trajectory planning.

The proposed high-level decision-making block was considered as a RL-Agent that seeks to find an optimal control policy $\pi : S \to A$ defined from the state space ($S$) to action space ($A$) that maximizes the total expected future rewards

$$R(\pi) = \mathbb{E}_\pi \left( \sum_t \gamma^t r(s_t, a_t) \right) \quad (1)$$

where $s_t$ and $a_t$ are state and action at time $t$, $r$ is the reward for $s_t$ and $a_t$, and $\gamma$ is the discount factor.

State-space involves the vectors of features such as velocities (both linear and angular) and lateral position of the ego vehicle, position and linear velocities of other cars represented in the ego vehicle frame. Action space includes three discrete actions (following, overtaking, and aborting), which trigger a different set of predefined parameters for the local planner as listed above. This action space was designed to accommodate an overtaking scenario on a two-lane road. We also take into account the ability of the agent to abort an overtaking action depending on additional observations that the vehicle gathers throughout the overtaking action.

The reward function was designed in such a way that the agent was rewarded at every step inversely proportional to its distance to the goal position determined by the global planner. The agent also gets a high reward once it finishes the episode in the correct lane. Additionally, accident cases are penalized with a huge negative reward and some small penalties were applied for frequent lane changing.

An optimal value function $Q^* : S \times A \to \mathbb{R}$ describing the expected cumulative rewards when starting from a given state and following the best policy is determined as

$$Q^*(s, a) = \max_\pi \mathbb{E} \left( \sum_{t=0} \gamma^t r(s_t, a_t) | s_0 = 0, a_0 = a \right) \quad (2)$$

Once the optimal value function ($Q^*$) is obtained, the optimal policy ($\pi^*$) that outputs the best action for a given state can be found as

$$\pi^*(s) = \underset{a}{\mathrm{argmax}} Q^*(s, a) \quad (3)$$

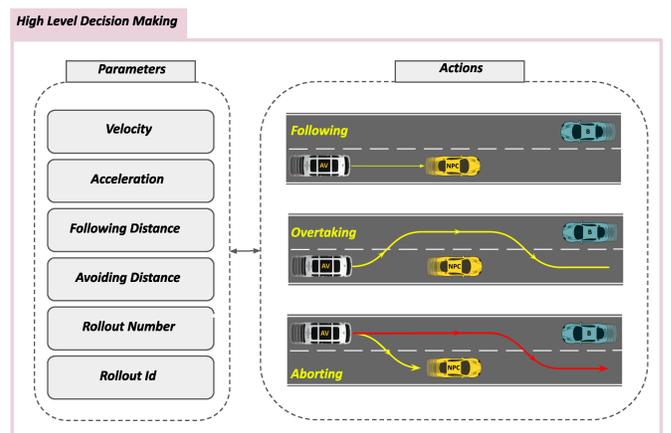

Fig. 2. Proposed learning based autonomous overtaking architecture

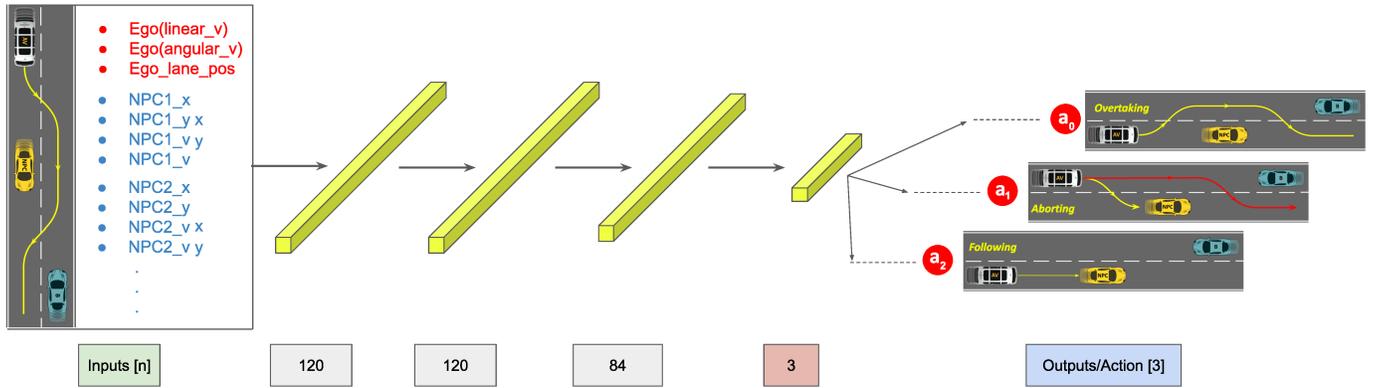

Fig. 3. Deep Q-Network employed in high-level decision making.

In order to approximate the value function of continuous state-space and discrete action-space, Deep Q-Networks (DQNs) [38] were employed, where $Q(s, a; \theta)$ is a deep neural network (Fig 10) and $\theta$ is the model parameters (weights and biases). DQN defines a second Q-network in exactly the same structure with parameters $\theta^-$, updates them only for every $C$ step with current parameters $\theta$, and holds fixed between individual updates to stabilize the learning process. The loss function for Q-learning update at every iteration is

$$L(\theta) = \mathbb{E}_{(s,a,r,s') \sim U(D)} \left[ \left( r + \gamma \max_{a'} Q(s', a'; \theta^-) - Q(s, a; \theta) \right)^2 \right] \quad (4)$$

where $s'$ and $a'$ are the next state and action. $D = (s, a, r, s')$ is a replay buffer that stores agent's experiences as tuples. To optimize the parameters $\theta$, loss function above is minimized on samples of experiences $(s, a, r, s') \sim U(D)$ uniformly drawn from the replay buffer.

### B. Openplanner as Integrated Planner

There are currently many open source planners that can be used as integrated planners. Nevertheless, some of open source integrated planners are difficult to use for a variety of reasons, such as the lack of documentation and the need for a specific platform. Based on the advantage of the Autoware framework [39], which is developed for autonomous driving applications, in the proposed method the Openplanner [33] is used as an Integrated planner.

In the OpenPlanner 2.0 [34], An integrated global planner uses a vector (road network) map to generate global reference paths. After generating a set of sampled roll-outs based on the global trajectory, the local planner generates an obstacle-free local trajectory. The optimal trajectory is determined by considering various factors, such as collisions, traffic rules, transitions, and distance from the center. Intentions and trajectories are estimated through an intention and trajectory estimator, while the behavior generator uses predefined traffic rules and sensor data to make predictions. A local planner is a set of tools that produces a smooth trajectory that can be tracked by path-following algorithms, such as Pure Pursuit [40]. A smooth trajectory created by the rollout generator can be replanned and selected by the behavior generator.

The openplanner algorithm was successfully implemented and tested as a mission and motion planning algorithm for AVs. Advantages of this algorithm include:
  (i) A framework for collaboration between different development teams, so they can share feedback and solve problems together.
 (ii) The software includes a complete set of plans (global planner, local planner, intention prediction and behavior planner)
(iii) Contains APIs for using the various planning modules in a Robot Operating System (ROS).
(iv) A broad range of platforms is supported, so the system can be used on many autonomous platforms. Support for both standard and open-source map formats, so that the system can be used with both.
 (vi) Provides a method to anticipate other vehicles' movements so that the system can plan accordingly.

In the Openplanner, lane change is supported by the local planning stages. In the first step, a reference plan is generated by the global planner, which includes global paths, lane change points, and backup reference paths. It is the task of the global planner to assign a lower cost to a target lane's path. In the next step, the trajectory generator produces smooth trajectories that are kinematically feasible for all possible paths. The next step is for the trajectory evaluator to determine the optimal roll-out trajectory in the chosen lane. Local planner push the autonomous vehicle to switch between lanes whenever possible because the target lane cost is always lower than other lanes. Finally, the behavior selector sends a replan signal to the global planner when the target lane cannot be reached. During the replan step, the global planner uses the backup path or finds an alternative path to the destination.

## IV. EXPERIMENTS

The experimental evaluation aims to answer the following questions:
- How does an RL agent learn high-level decisions for successful overtaking?
- Does learning-based high-level decision-making improve safety?
- What is the advantage of learned high-level decision-making over rule-based decision-making?

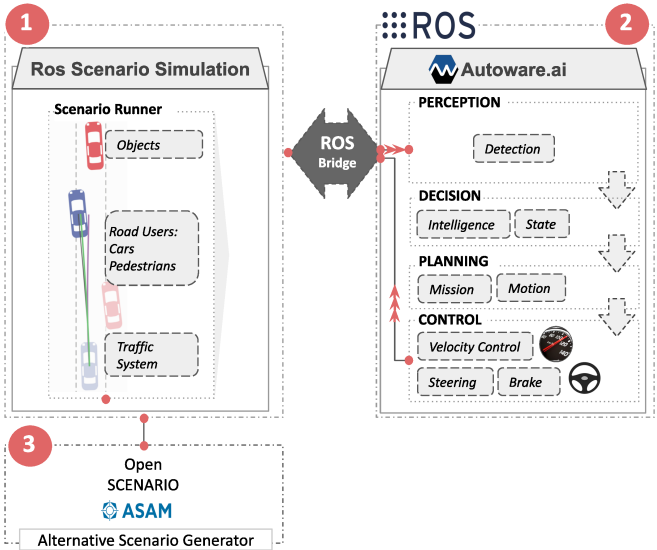

Fig. 4. Overview of the Ros scenario simulation architecture. Scenario generator (1). Scenario runner (2). ROS (3).

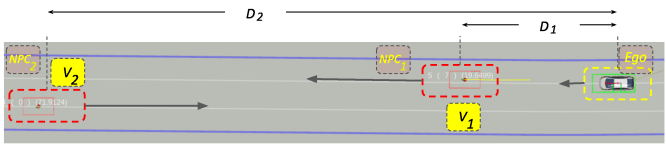

Fig. 5. Different parameters for creating 1000 random scenarios. $D_1$ and $D_2$ are distance between Ego vehicle, $NPC_1$ and $NPC_2$ respectively. $V_1$, and $V_2$ are velocity of $NPC_1$ and $NPC_2$ respectively.

- How does high-level decision-making merge with integrated planning?

In order to answer these questions, first, the RL Agent is trained by using the ROS scenario simulator [41]. In the next step, to evaluate the abortable overtaking algorithm in the simulation environment, the algorithm was implemented in high-fidelity software with the simulator in the loop (SVL simulator [42]). And finally, the trained RL-Agent was used in the experiment with an automated shuttle on the testbed.

### A. Training RL-Agent with the ROS Scenario Simulator

To prevent unsafe conditions during the collection, evaluation, and deployment of data, it is crucial to choose a simulation engine that takes into account the complexities of real-world situations. A simulator must meet certain requirements that will influence the simulation development, including functionality, testing, and integration of machine learning (ML). An overview of the training workflow, including ROS scenario simulator, Autoware stack, and scenario generator, is presented in Fig. 4.

ROS Scenario simulator (annotated 1 in Fig. 4) is an open source framework for evaluating path planning algorithms in autonomous driving. The simulator is based on CommonRoad [43] and ASAM Openscenario [44]. The simulator allows controlling the ego vehicle in simulation using the AV software running on a physical vehicle. As AV framework, we use Autoware (annotate 2 in Fig. 4), an open-source ROS-based autonomous driving stack, which was connected to the simulator using a ROS bridge. The bridge allows two-way communication between the simulator and Autoware, transmitting simulated sensor data from simulation to Autoware and vehicle control commands from Autoware to the simulator. The simulation results are recorded to allow analysis. To generate and simulate the variety of scenarios occuring in overtaking, we used a scenario generator based on the ASAM OpenScenario library (annotate 3 in Fig. 4). The generated dataset includes 1000 random scenarios, with randomized number of non- player characters (NPCs), velocities of NPCs, and relative starting positions of the vehicles, as illustrated in Fig. 5. The generated scenarios consider overtaking an NPC on a two-lane road with another NPC on the opposite lane. We trained the RL-Agent for 8000 iterations. Evaluation is carried out by the trained Rl-Agent in the next section.

### B. Evaluating the RL-Agent with the ROS Scenario Simulator

To evaluate the performance of the trained RL-Agent, we compared its performance to OpenPlanner in 3000 scenarios. Results of the comparison are shown in Table I. For the evaluation process, 3000 scenarios with a range of velocity for NPC1 and NPC2 ( from 1 to 3 m/s) and different position were created by scenario generator API. The RL-Agent and Openplanner were evaluated with these scenarios. As seen in TABLE I the successfull overtaking range with RL-Agent is 92.2% in compare of 52.7% by Openplanner. Based on the results in the TABLE I, the completion time of successfull scenarios in the RL-Agent is 27.9 seconds in compare of 28.6 seconds with Openplanner shows that the proposed method performs fast decisions between overtaking and aborting. To further evaluate the performance of proposed RL-Agent, we analyzed crashes in details in the TABLE I.In compare of the crash with the NPC1 or NPC2 , with the RL-Agent 82.5% of crashes happend with the NPC1. On the other hand 83.5% of crashes with Openplanner occured on the opposite lane with NPC2. According to these results, the abort feature in the proposed RL-Agent increases safety by preventing collisions with oncoming vehicles. As seen in TABLE I, with 3000 scenario only there is one scenario which failed on both methods. The velocity range for the NPC1 and NPC2 was defined between 1 to 3 m/s (3.6 to 10.8 km/h). However most of the crashes happened with average speed 2.0 - 2.1 m/s.

### C. Evaluating the RL-Agent with SVL simulation

The iseAuto control software is compatible with multiple realistic car simulators controlled by a physics engine, including SVL and CARLA [45]. Modern game engines like Unreal and Unity give them the ability to create complex virtual environments. Before evaluation in the real world, an assessment of the trained RL-Agent for overtaking is carried out using the SVL simulator. In this simulator, environments, as well as car models, are provided by the Unity game engine. To test the RL-Agent in the simulator, a detailed iseAuto 3D model was implemented inside Unity and assigned to the engine [46].

TABLE I
COMPARISON RESULTS OF OVERTAKING MANEUVER BETWEEN THE PROPOSED METHOD AND OPENPLANNER IN 3000 SCENARIOS.

|  | RL-Agent | OP only |
|---|---|---|
| Successful overtaking | 92.2% | 52.7 % |
| Completion Time of Successful Scenarios (s) | 27.9 | 28.6 |
| Crash with NPC1 | 82.5% | 16 % |
| Crash with NPC2 | 17.4% | 83.5 % |
| Crash in same scenarios | 3.3% | 3.3% |
| Average speed of NPC1 in failed scenarios(m/s) | 2.1 | 2.0 |
| Average speed of NPC2 in failed scenarios(m/s) | 2.1 | 2.1 |

During the evaluation, three NPCs were placed on a two-lane road, and the RL-Agent capabilities were assessed using an overtaking scenario created within the SVL simulator. Fig 6 illustrates the abortable overtaking in four frames with a ROS visualization screen showing the point cloud of the simulated environment, the shuttle, and the vehicle's trajectory. In Fig 6(1), the automated shuttle travels in the "following" mode. In the "following" mode, the "overtaking" is disabled by the RL-Agent. So the trajectory planning has permission to use the middle rollout. When the automated shuttle detects an NPC driving at a slower speed on the path, RL-Agent changes the mode to "overtaking" mode (Fig 6(2)). During the overtaking, the automated shuttle detects another NPC on the opposite lane (Fig 6(3)) so the RL-Agent aborts the overtaking and switches to the "following" mode. In the final frame, when the automated shuttle detects that the opposite lane is free, the RL-Agent changes the mode to overtaking and overtakes the front vehicle (Fig 6(4)). Fig 9 shows the steering data for the simulation in the SVL simulator. As seen in Fig 9, the proposed method aborts the overtaking smoothly at distances 20 to 30 meters and completes the overtaking at distances 60 to 90 meters successfully. In addition, the proposed method improves high-level decision-making with the same low-level controllers. In this way, performance is maintained and unexpected failures are avoided. According to the results of this experiment, the experimental setup is reliable with regard to the simulation environment. As a result, the proposed simulation shows that it is not necessary to directly test newly developed algorithms on AVs. Instead, these algorithms can be first evaluated in the simulation platform. While the selected scenarios did not include all the real-world complexity, they were an initial step towards studying and introducing the verification platform. It is possible to create and test scenarios such as multiple vehicles overtaking at high and dynamic speeds coming from the back or in the opposite direction in a simulation without any inherent danger.

*D. Setup physical experiments on the real Autonomous shuttle*

Our study was conducted at TalTech, Estonia, using the iseAuto. The iseAuto is an automated shuttle operated on the campus for experimental and educational purposes by the AVs research group (see Fig 7). The objective of the iseAuto project was to build an open-source automated shuttle and build a smart city test bed at TalTech. It allows researchers to build a variety of urban mobility projects for the future. A digital twin of the automated shuttle and the testbed allows designers to simulate all development prior to implementing it. The simulation environment, interface, and concept described in detail in [47], [48].

ROS (Robotic Operating System) provides the high-level software architecture of the iseAuto platform. Autoware [49], an open-source ROS-based autonomous driving software, is used for perception, detection, and planning. It already includes many advanced algorithms, such as lane following, obstacle avoidance, traffic light detection, and lane detection. Localization and path following are performed using Lidars and global navigation satellite systems (GNSS). Velodyne

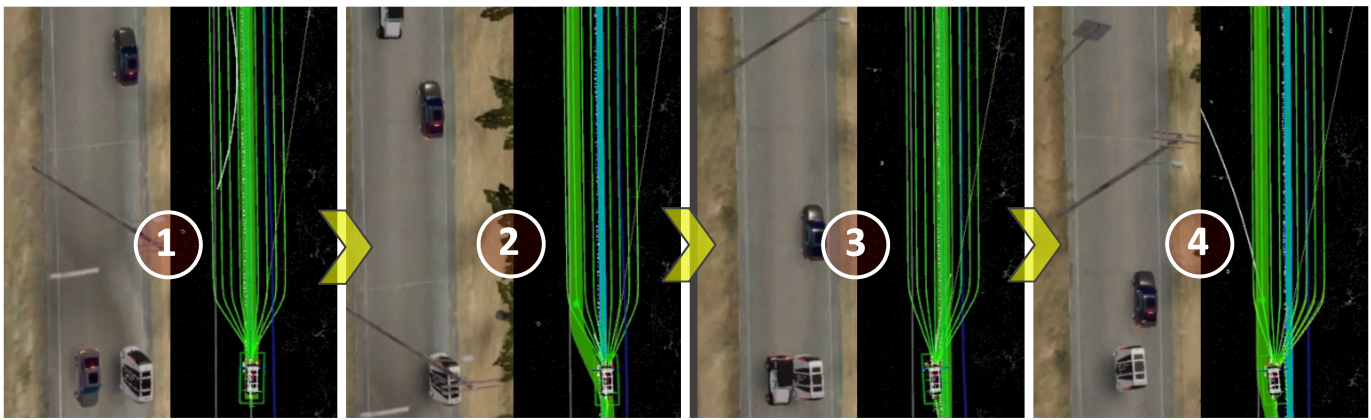

Fig. 6. Simulation of an abortable overtaking scenario in the SVL simulator with ROS visualization.1) Following. 2) overtaking. 3) aborting the overtaking. 4) overtaking.

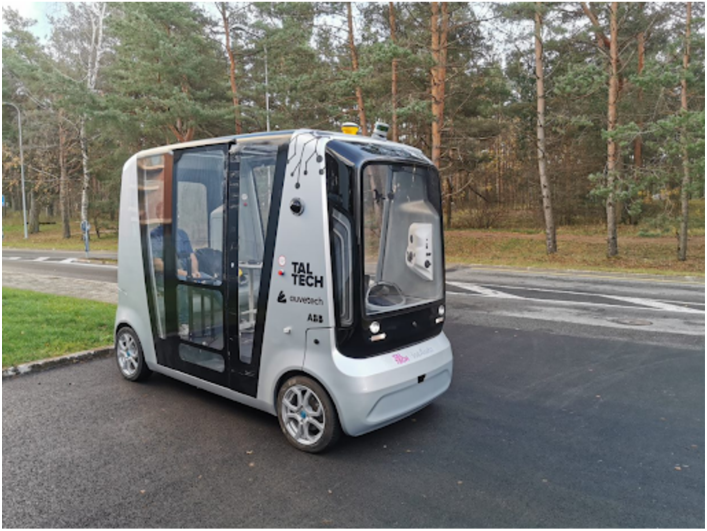
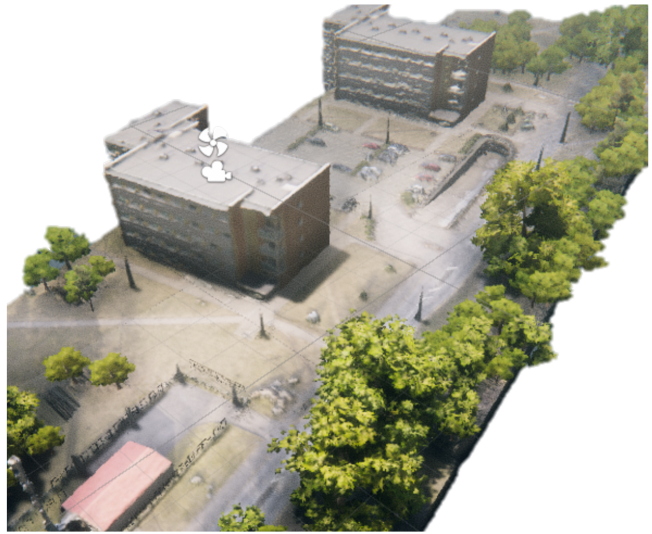

Fig. 7. An image of the TalTech iseAuto autonomous shuttle (left) and an eagle view of the test site on the TalTech campus (right).

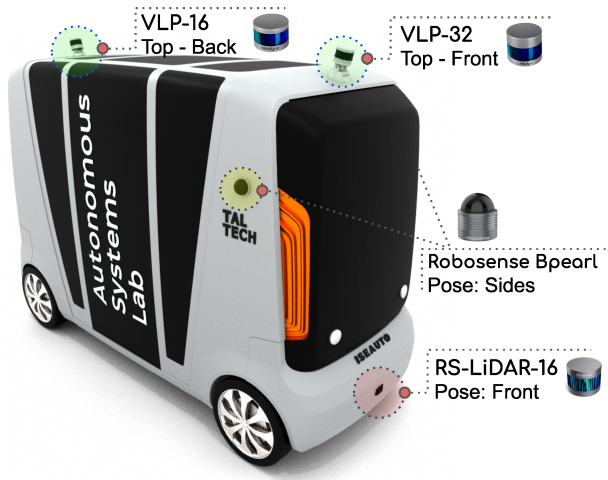

Fig. 8. Various types of Lidars are provided on the iseAuto shuttle bus with an indication of their position.

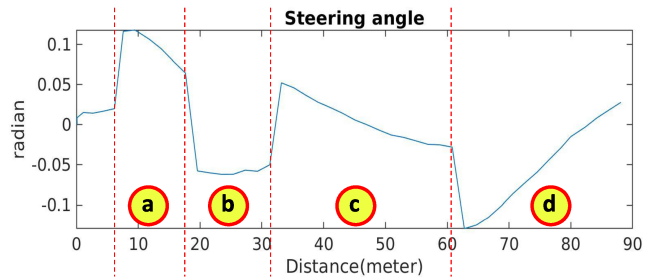

Fig. 9. Steering angle changes for abortable overtaking with trained RL-Agent in SVL :a) AV starts overtaking; b) Av aborts the overtaking because of oncoming NPC; c) Av resumes unfinished overtaking; d) AV completes the overtaking.

Lidar (VLP-32) and (VLP-16) are located at the front and rear of the vehicle, and two Robosense RS-Bpearls are mounted on the front sides of the vehicle to decrease blind spots. Additionally, an RS-Lidars-16 is installed on the front bumper to detect objects not visible to the other Lidars. Lidar sensors are shown in Fig 8.

Based on the simulation results provided in Section IV, the parameters that conducted the best performance with the proposed method were applied to the iseAuto for use in AV testing at TalTech. There are two GeForce GTX-1080-Ti GPUs in the main processing unit of the shuttle, with an AMD Ryzen threadripper 1950x processor and an Ubuntu 18.04 operating system and the shuttle speed was limited to 15 km/h.

e proposed method aborts the overtaking smoothly at distances 20 to 30 meters and completes the overtaking at distances 60 to 90 meters successfully.

## V. Discussion

To interpret the learned policy, we illustrate it in Fig. 10 which shows the effect of vehicle positions on the decision for particular velocities. The colors indicate decisions. Three states denoted by A, B, and C are illustrated on the left side of the figure, as an example of using the decision map to evaluate the policy by an expert. In state A, the overtaking is aborted (action 1) since the vehicle on the opposite lane is getting closer. In state B, vehicle following (action 2) is chosen because the vehicle in fron of the ego vehicle is still far away and there is another vehicle on the opposite lane. In state C, overtaking is performed since the other vehicle on the opposite lane (NPC2) is sufficiently far away and the ego vehicle is sufficiently close to the front vehicle (NPC1). The evaluation shows that the learned policy can be manually inspected by an expert to study its behavior and safety. A similar investigation could be made for other input variables to further study these factors before deployment.

The performance of the maneuvers was assessed using extensive experiments conducted with the iseAuto automated shuttle. As part of the simulation and experiment setup, the steering angle feedback was recorded. Based on the simulation and experimental results, the overtaking maneuver with an aborting ability guarantees safety, efficiency, and high reliability during the operation.

### A. Validation in real world

The trained RL-Agent was used to run long experiments while recording the data, similar to the simulation environment. Three different case studies have been presented in an experimental setup to further demonstrate the effectiveness of the proposed method. A drone was used to record the trajectory driven by the vehicle from the top view.

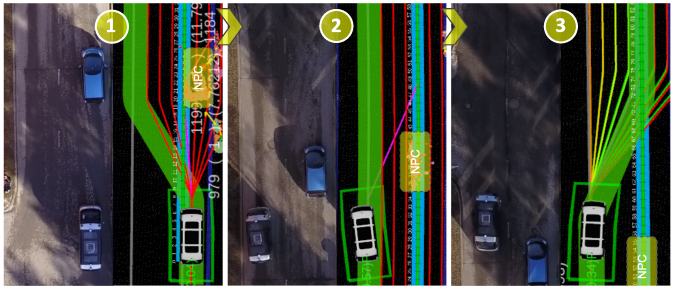

Fig. 11. overtaking from one vehicle in real experiment with ROS visualization.1) lane change for overtaking. 2) passing the vehicle. c) lane change back to global path.

*1) Overtaking on the road from one vehicle:* Similar to the simulation scenario, this experiment evaluates the overtaking using the trained RL-Agent from one vehicle on the road. This experiment is described in three stages (Fig 11). In each stage, the left frame is the top view captured by a drone and the right frame is a screenshot of the ROS visualization software that displays the current iseAuto position and the global and local path. Fig 11 (1) shows that the iseAuto detects the NPC and correctly creates a local path for overtaking. In the next stage the iseAuto continues traveling in the opposite lane to pass the NPC (Fig 11 (2)). In the final stage, the iseAuto passes the NPC and starts to return to the global path (Fig 11 (3)).

*2) Overtaking on the road from one simulated car with two other simulated cars in the opposite lane :* This experiment is a test of the RL-Agent's ability to overtake an NPC with two NPCs on the opposite lane. The isAuto is used for this experiment. To ensure the safety of all participants, three simulated NPCs were placed on the road at varying speeds and positions (see Fig 12). Steering data were recorded throughout the experiment. As seen in Fig 13, the AV starts overtaking. Section (a) shows that the iseAuto aborts the overtaking when it detects an NPC2 in the opposite lane. Immediately after the NPC2 passed by, the iseAuto starts the overtaking. There is a sharp abort observed in section (b). An actual car appeared on the opposite lane during the second overtaking try and the iseAuto aborted the overtaking. Finally the iseAuto finished overtaking successfully.

*3) Overtaking on the road from one car with other car in the opposite lane :* In this experiment, two real cars were used in a similar way to the simulation (see Fig 14 and Fig 15). Each picture shows two different shots, one from real life and the other from the ROS visualization software that demonstrates the aborting process in the overtaking maneuver. The autonomous shuttle correctly detected NPC2 and aborted the operation. The experiments were carried out twice, and in each try, NPC2 was driven at a different speed. On the first attempt, NPC2 traveled at a speed of 10 km/h. According to Fig 14, during the aborting, there was a distance of 20 meters between iseAuto and NPC2. On the second try, the speed of NPC2 was increased to 20 km/h and as can be seen in Fig 15, the distance between the iseAuto and NPC2 reached 8.1 meters. Fig 16 shows the driving path of the experiment. On both tries, the AV shuttle aborted the overtaking when there was a vehicle in the opposite lane and finished overtaking successfully when there was no traffic in that lane.

### VI. CONCLUSION

In this paper, a learning-based abortable overtaking maneuver for an autonomous shuttle is experimentally studied. To overcome the limitations of the current state-of-the-art algorithms, a high-level decision-making approach based on reinforcement learning is proposed to be added to the integrated planner. In our design process, safety and reliability were the most important criteria for automated vehicle overtaking. A high-fidelity simulation method was used to validate the efficiency of the algorithm and predict its behavior. Real-world experiments were conducted on an automated shuttle to explore the performance of the proposed abortable overtaking algorithm. Based on the results, the proposed method improved safety and prevented unsafe overtaking. A further study should focus on boosting the safety from 92.2% to 100% in TABLE I. The present study provides a platform for the development of more advanced overtaking methods for complex scenarios, including driving vehicles with abnormal behavior, extending and improving perception for quick and precise detection, and overtaking at high speeds.

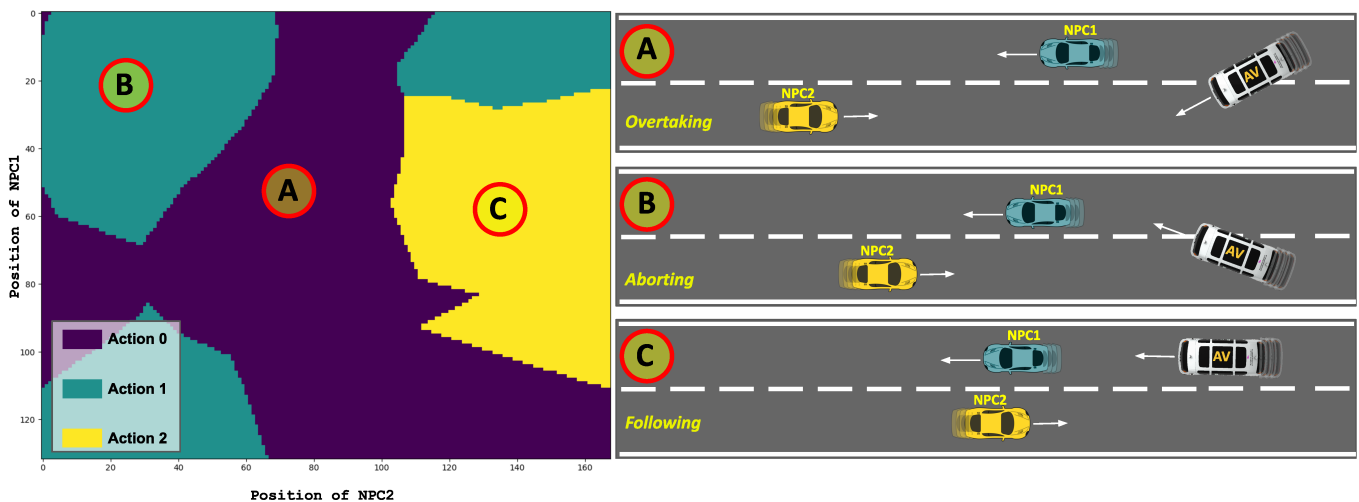

Fig. 10. Interpretability of the proposed DQN-based high-level decision-making mechanism for overtaking.

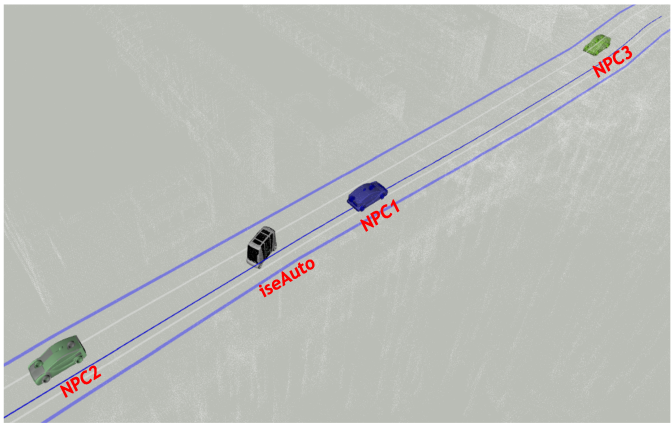

Fig. 12. Overtaking on the road from one simulated car with two other simulated cars in the opposite lane.

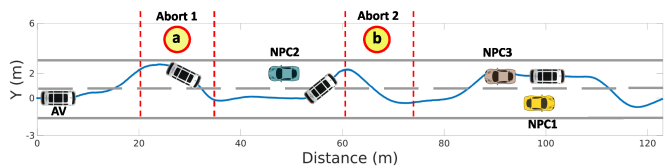

Fig. 13. iseAuto driving path in the first experiment; a) aborting the overtaking for oncoming NPC2. b) aborting the overtaking for oncoming NPC3

## Declaration of conflicting interests

This article's author(s) reported no potential conflicts of interest in relation to its research, authorship, or publication.

## Acknowledgement

The author(s) disclosed receipt of the following financial support for the research, authorship, and/or publication of this article: This research has received funding from two grants: the European Union's Horizon 2020 Research and Innovation Programme, under the grant agreement No. 856602, and the European Regional Development Fund, co-funded by the Estonian Ministry of Education and Research, under grant agreement No 2014-2020.4.01.20-0289.

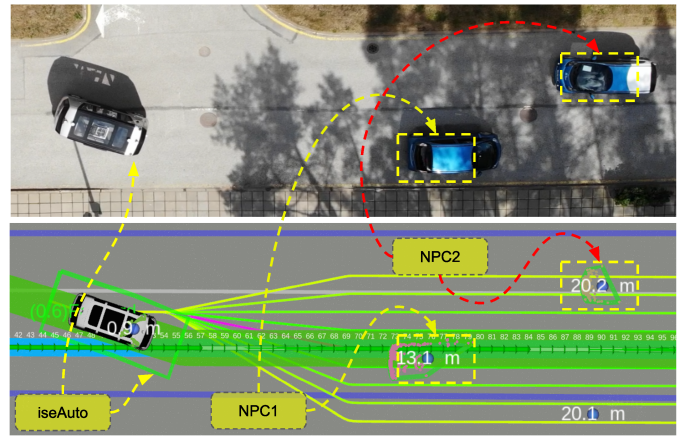

Fig. 14. The first attempt of the experiment with two real vehicles on the road (drone view on top, Ros visualization on bottom).

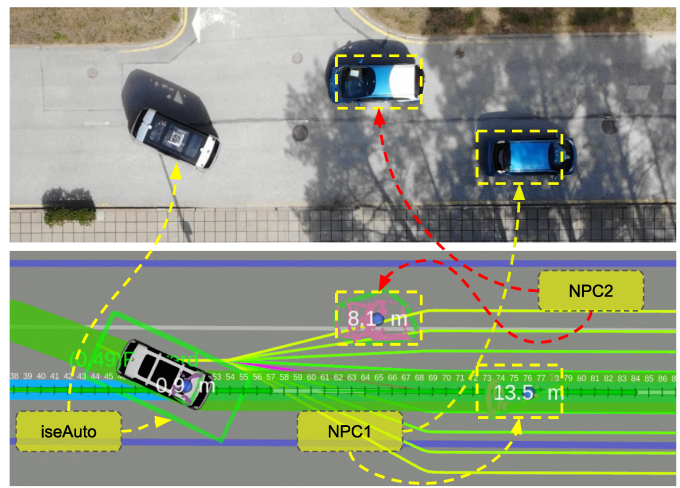

Fig. 15. The second attempt of the experiment with two real vehicles on the road (drone view on top, Ros visualization on bottom).

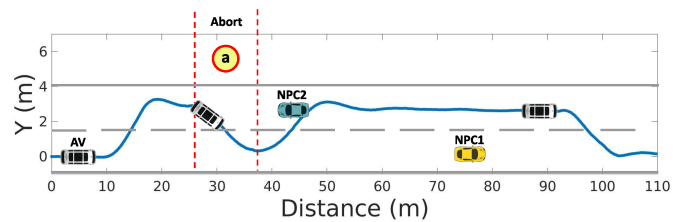

Fig. 16. iseAuto driving path according to its position during the third experiment; a) aborting the overtaking for oncoming NPC2.

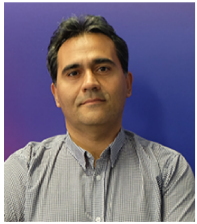

**Ehsan Malayjerdi** received his Master's degree in Artificial Intelligence at IAUM. He is currently studying for a PhD in the Department of Mechanical and Industrial Engineering at Tallinn University of Technology. He has authored several papers on the topics of human-robot interaction, path and trajectory planning, and control for autonomous vehicles. Currently, his research focuses on autonomous vehicle developments. The focus of his current research is developing the mission and motion planning of a real AV shuttle.

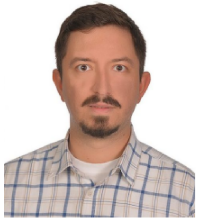

**Gokhan Alcan,** received his B.Sc., M.Sc. and Ph.D. degree in Mechatronics Engineering from Sabanci University, Istanbul, Turkey in 2013, 2015 and 2019, respectively. Since 2020, he has been working as a postdoctoral researcher in Intelligent Robotics Research Group led by Prof. Ville Kyrki in Aalto University, Finland. His primary research insterest include safe model predictive control, safe robot learning and decision making for autonomous driving.

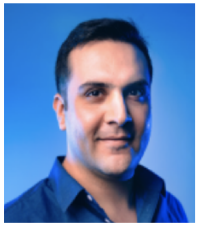

**Eshagh Kargar** received the B.Sc. and M.Sc. degrees in electrical engineering from University of Tehran, Iran, in 2013 and 2016, respectively.

In 2019, he joined the Intelligent Robotics Group at Aalto University, Helsinki, Finland, where he is a Doctoral Candidate. His primary research interests include robotic perception, decision making, and learning

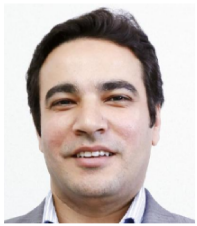

**Hatem Darweesh, Ph.D. (Member, IEEE)** received the B.Sc. and M.Sc. degrees from Ain Shams University, Egypt, in 2002 and 2009, respectively, and the Ph.D. degree from the Graduate School of Informatics, Nagoya University, Japan, in 2020. From 2002 to 2013, he worked as a Teaching Assistant at Modern Academy, Maadi, Egypt. In 2013, he started working at ZMP Inc., Japan, as an Autonomous Driving Team Leader. In 2016, he enrolled in a Ph.D. Program at the Graduate School of Informatics, Nagoya University. Since 2020, he has been working as a Postdoctoral Researcher at Nagoya University. He is currently the CEO at ZATiTECH Inc., Nagoya. His main research interests include planning and control of autonomous vehicles, mapping, and simulation.

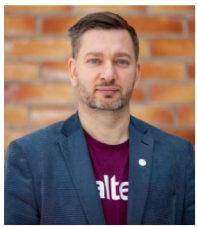

**Raivo Sell, Ph.D. (Member, IEEE)** received his Ph.D. degree in Product Development from Tallinn University of Technology in 2007 and currently working as a professor of robotics at TalTech. His research interest covers mobile robotics and self-driving vehicles, smart city, and early design issues of mechatronic system design. He is running the Autonomous Vehicles research group at TalTech as a research group leader with a strong experience and research background in mobile robotics and self-driving vehicles. Raivo Sell has been a visiting researcher at ETH Zürich, Aalto University, and most recently at Florida Polytechnic University in the US, awarded as a Chart Engineer and International Engineering Educator. Dr. Sell is a member of IEEE Robotics and Automation Society and Estonia section.

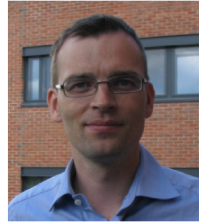

**Ville Kyrki (M'03–SM'13)** received the M.Sc. and Ph.D. degrees in computer science from Lappeenranta University of Technology, Lappeenranta, Finland, in 1999 and 2002, respectively.

In 2003-2004, he was a Postdoctoral Fellow with the Royal Institute of Technology, Stockholm, Sweden, after which he returned to Lappeenranta University of Technology, holding various positions in 2003-2009. During 2009-2012, he was a Professor of computer science, Lappeenranta University of Technology. Since 2012, he is currently an Associate Professor of Intelligent Mobile Machines with Aalto University, Helsinki, Finland. His primary research interests include robotic perception, decision making, and learning.

Dr. Kyrki is a Fellow of Academy of Engineering Sciences (Finland), and a member of Finnish Robotics Society and Finnish Society of Automation. He was a Chair and Vice Chair of the IEEE Finland Section Jt. Chapter of CS, RA, and SMC Societies in 2012-2015 and 2015-2016, respectively, Treasurer of IEEE Finland Section in 2012-2013, and Co-Chair of IEEE RAS TC in Computer and Robot Vision in 2009-2013. He was an Associate Editor for the IEEE Transactions on Robotics in 2014-2017.